# 基于视觉识别的智能寻路车辆设计

作者：郭湛予，郭珅元，王珈珑，冯伊凡

## 摘要

智能送药小车是一种先进的智能送药设备，与传统的人力送药相比，具备更高的送药效率以及更低的失误率。

本项目中，设计并制作了一辆智能送药小车，可通过视觉识别技术来实现道路路线的识别以及目标病房房间号的识别。小车根据识别到的房间号来选择对应的路线，并将药品准确运送到目标病房，在药品送达后可自行返回药房。该智能送药小车使用直流电源供电，由电机驱动模块控制两个直流电机，克服了转弯转角过大的偏离问题。小车巡线功能使用闭环控制来提高巡线准确性和小车速度的可控性。病房号码的识别通过带微控制器的摄像模组来完成，并且具备环境亮度自适应调节、畸变校正、自动标定等功能。两协同送药小车间的通信由蓝牙模块实现，达到了高效准确的通信与交互。

实验表明，该智能送药小车能准确识别房间号码并规划路线，将药品送达远、中、近端病房，并且具有速度快、判断准确的特点。除此之外，两辆送药小车可协同送药至同一病房，效率高且配合度高。

**关键词**：智能送药小车；PID 控制；视觉识别；图像处理

# Abstract

Intelligent drug delivery trolley is an advanced intelligent drug delivery equipment. Compared with traditional manual drug delivery, it has higher drug delivery efficiency and lower error rate.

In this project, an intelligent drug delivery car is designed and manufactured, which can recognize the road route and the room number of the target ward through visual recognition technology. The trolley selects the corresponding route according to the identified room number, accurately transports the drugs to the target ward, and can return to the pharmacy after the drugs are delivered. The intelligent drug delivery car uses DC power supply, and the motor drive module controls two DC motors, which overcomes the problem of excessive deviation of turning angle. The trolley line inspection function uses closed-loop control to improve the accuracy of line inspection and the controllability of trolley speed. The identification of ward number is completed by the camera module with microcontroller, and has the functions of adaptive adjustment of ambient brightness, distortion correction, automatic calibration and so on. The communication between two cooperative drug delivery vehicles is realized by Bluetooth module, which achieves efficient and accurate communication and interaction.

Experiments show that the intelligent drug delivery car can accurately identify the room number and plan the route to deliver drugs to the far, middle and near wards, and has the characteristics of fast speed and accurate judgment. In addition, two drug delivery trolleys can cooperate to deliver drugs to the same ward, with high efficiency and high cooperation.

# 目录

# 1 方案设计与论证

## 1.1 总体方案描述

该系统主要利用运动控制原理，实现小车路径的精确控制，同时结合机器视觉，实现送药小车自主规划并识别房间号的功能。

图 1-1 是送药小车系统的构成框图，主要由控制面板，接触开关，控制模块，信号采集，通信模块构成。

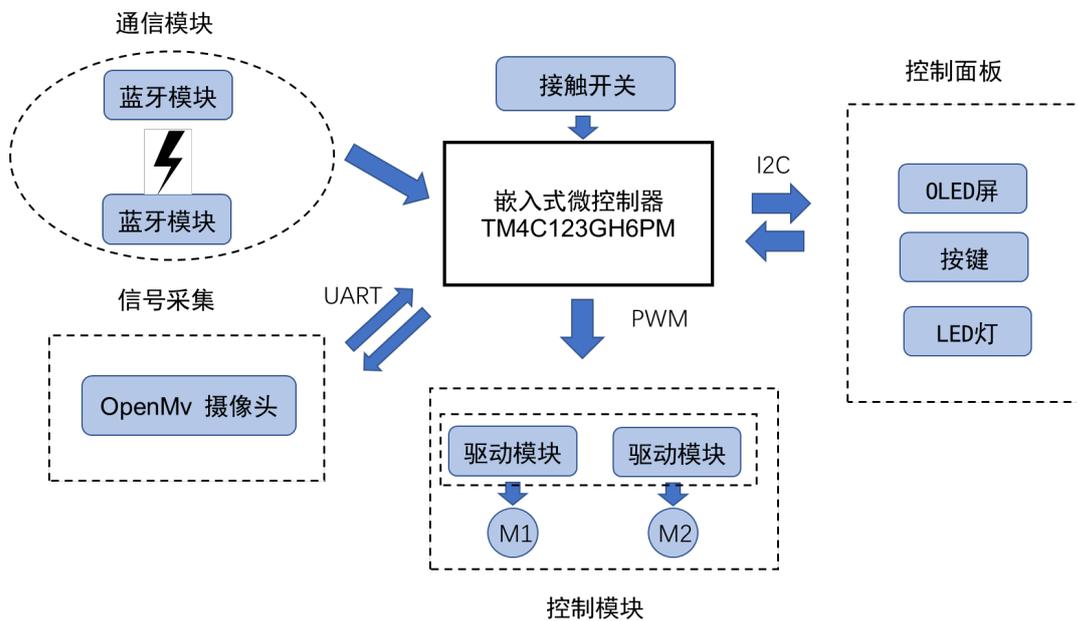

图 1-1 智能送药小车系统的构成框图

具体的工作步骤为：

步骤一：OpenMv 摄像头对所需房间号进行识别。

步骤二：通过控制面板上的按键启动小车。

步骤三：砝码放置后触发接触开关，控制驱动模块驱动小车前进。

步骤四：OpenMv 与单片机使用串口总线将所识别到的中线的 x 坐标以及房间号码的 x，y，z 坐标发送到单片机上。

步骤五：单片机读取中线的 x 坐标，根据计算出的中线坐标以及房间号码坐标相对于小车中心的偏差，控制对电机输出的 PWM 波大小。

步骤六：单片机读取地面上房间号码的 x，y，z 坐标，判断与小车的相对位置后选择合适的路线。

步骤七：到达指定病房后，触发控制面板上的红色 LED 灯亮起。砝码卸下触发接触开关断开，LED 指示灯熄灭。

步骤八：单片机接收接触开关的信号，再次启动电机，小车返回药房区域。

## 1.2 系统结构

### 1.2.1 系统机械结构

图 1-2 是系统的整体机械结构设计图，该系统主要包括小车底座、电机驱动板、万向轮、直流无刷电机以及 OpenMv 图像识别模块五部分。

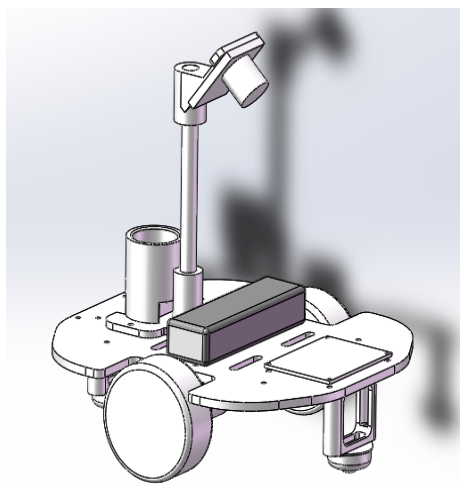 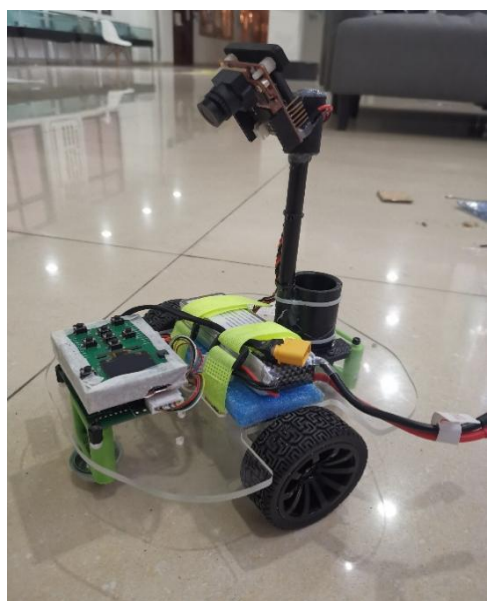

图 1-2a 系统机械结构　　　　图 1-2b 实物图

图 1-2 系统整体模型

横向放置的直流无刷电机连接车轮作为动力轮，通过输入 PWM 脉冲信号控制小车前进、后退、转弯以及停止等功能，纵向放置的万向轮作为支撑轮保持小车平衡。

为了提高摄像头对引导标识识别的准确度，控制 OpenMv 摄像头对地角度约为 45°，靠近小车中心，可恰好识别到小车前端数字。

### 1.2.2 系统硬件结构

执行硬件结构 3D 模型图如下，

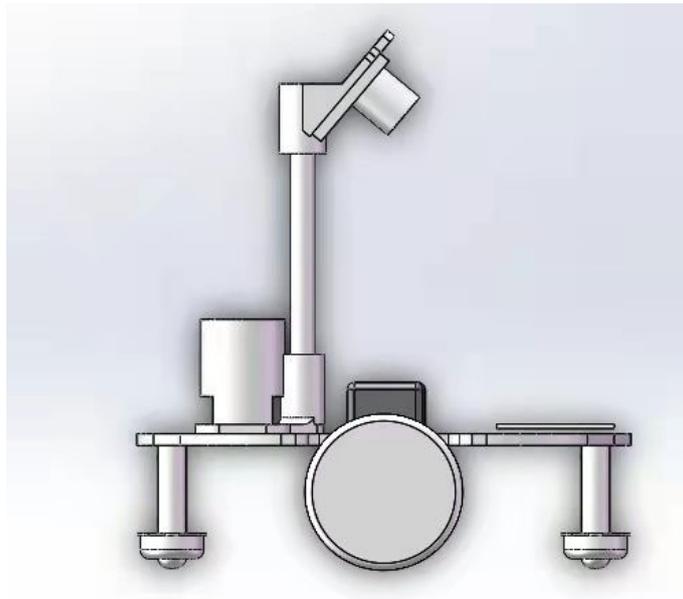

图 1-3 小车 3D 模型

## 1.3 可行性分析

本作品由主控板、含万向轮两驱小车、摄像头、蓝牙、电池等模块构成。红色线循迹与数字识别是本次比赛的核心难点，由 OPENMV 实施，循迹功能由开源既有库完成，数字识别采用模板匹配，基于逻辑设计整合两大功能，集成于摄像头，该方案用软件复杂度代替双摄像头的硬件复杂度，从测试效果来看，工作量大，但可实现题目要求。

受 30cm 赛道限制，设计带万向轮的两驱方案，相比于四轮方案，转弯半径小（可原地旋转），车体长度有所减小，手动测试表明，设计方案满足赛题要求，且灵活度高。

# 2 系统软硬件设计

## 2.1 执行机构控制算法与驱动

### 2.1.1 PID 控制算法

循迹部分采用在工业控制中运用广泛的 PID 控制器,循迹控制系统采用位置式 PID 算法。根据采样时刻的偏差来计算控制量， PID 控制公式表达式：

$$u_K = K_p[e_K + \frac{T}{T_i}\sum_{j=0}^{K} e_j + T_d \frac{e_K - e_{K-1}}{T}]$$

或

$$u_K = K_p e_K + K_i \sum_{j=0}^{K} e_j + K_d(e_K - e_{K-1})]$$

单片机将 OpenMv 发送回的数据进行 PID 运算,计算 PWM 方波控制信号的占空比,改变方波信号的频率和方波数量,可控制直流无刷电机的运动速度和旋转方向,实现循迹功能。

### 2.1.2 直流电机驱动电路

使用直流电机驱动板,脉冲波形的频率控制电机转动速度和方向,脉冲波形的脉冲数控制步进电机的转动角度和速度,从而控制直流电机正转、反转以及停止。

## 2.2 图像处理算法

采用轮廓识别的算法,可以通过 OpenMv 内置的摄像头模组将数字 1-8 进行采样,其中数字 1 和 2 只采取一个轮廓识别的样本,数字 3-8,每个数字需要采样 4 次,分别放置在不同的识别位置,来满足在 4 个数字中摘取题目中所需要的数字,取样轮廓如下图所示。

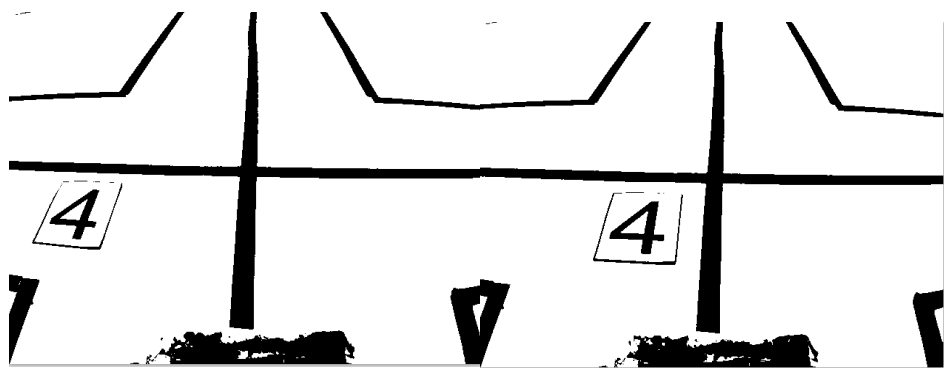

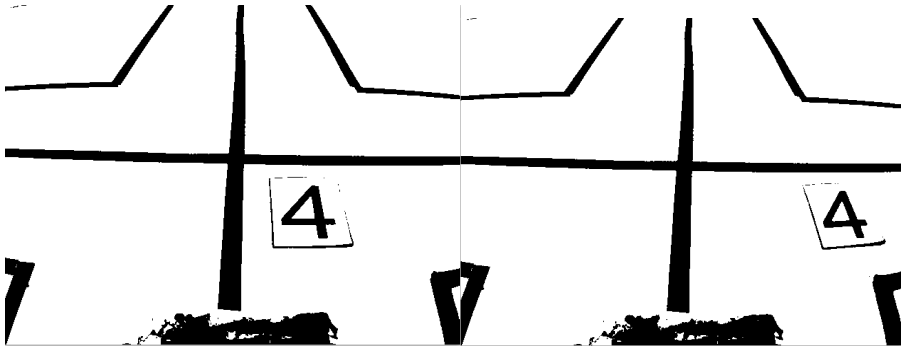

图 2-1 轮廓识别的算法图

图像处理使用 OpenMv 摄像头模块，图像处理具备环境亮度自适应调节、畸变校正、自动标定等功能。图像处理算法流程框图见附录 7.4。

## 2.3 程序流程

根据任务要求设计程序框架。程序开始时进行系统初始化，设置系统时钟中断，按键初始化以及选择执行的任务，将数字卡片放置在车前进行数字选择。

## 2.4 数字识别及循迹电路

启动前出示数字卡片设置寻找的数字；压触开关模拟药品是否放置；两个直流无刷电机为转动轮提供动力；OpenMv 对图像进行处理返回线路中心坐标完成循迹工作。电路如图 2-2 所示。

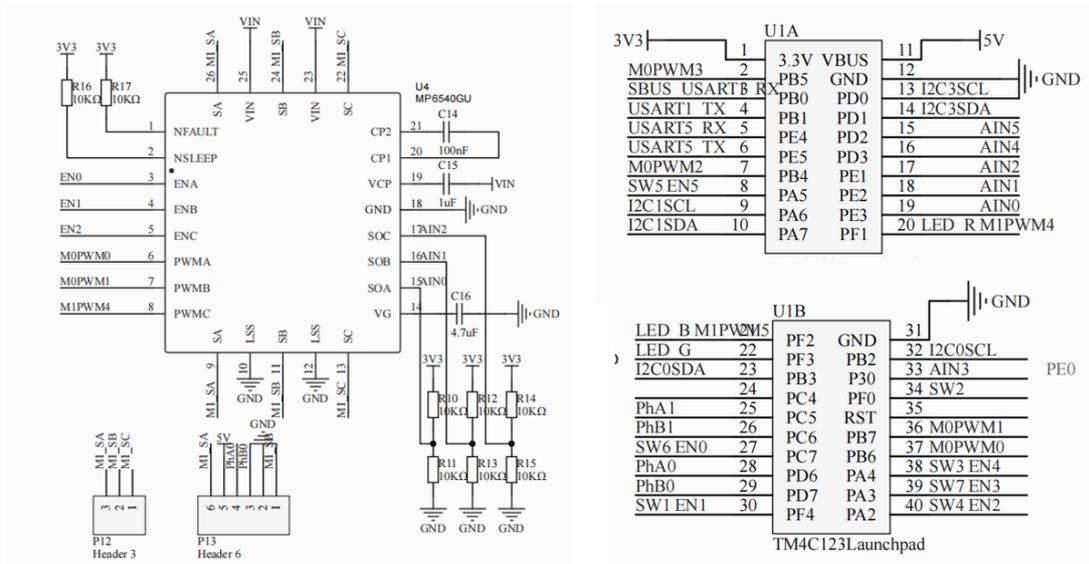

图 2-2 电机驱动模块电路图

# 3 系统测试

## 3.1 测试仪器

在室温条件下，测试所用仪器如表 3-1 所示。

表 3-1 测试仪器

| 编号 | 名称 | 型号 |
|---|---|---|
| 1 | 数字卡片 | 无 |
| 2 | 200g 砝码 | 无 |

## 3.2 指标测试结果

### 3.2.1 基础要求

（1）近端送药循迹

| 病房号 | 数字是否识别成功 | 是否送达 |
|---|---|---|
| 1 | 成功 | 送达 |
| 2 | 成功 | 送达 |

（2）中端送药循迹

| 病房号 | 数字是否识别成功 | 是否送达 |
|---|---|---|
| 3 | 成功 | 送达 |
| 4 | 成功 | 送达 |

（3）远端送药循迹

| 病房号 | 数字是否识别成功 | 是否送达 |
|---|---|---|
| 5 | 成功 | 送达 |
| 6 | 成功 | 送达 |
| 7 | 成功 | 送达 |
| 8 | 成功 | 送达 |

### 3.2.2 发挥部分

（1）两个小车协同运送药品到同一指定中部病房。小车 1 识别病房号装载药品后开始运送，到达病房后等待卸载药品；然后，小车 2 识别病房号装载

药 品后启动运送，到达自选暂停点后暂停，点亮黄色指示灯，等待小车 1 卸载；小车 1 卸载药品，开始返回，同时控制小车 2 熄灭黄色指示灯并继续运送。总时间约为 35 秒。

## 3.3 误差分析

经分析，系统的主要误差来自：场地亮度、反光，风速，水平面的平整程度，灰尘对车轮摩擦力的影响等。

# 5 总结

寻迹小车以 TivaTM4C123FH6PM 为主控制器，以 OpenMv 为图像处理器，TivaTM4C 根据 OpenMv 对引导标识的位置信息进行 PID 运算，输出 PWM 波作用于电机以达到追踪引导标识的目的。因为循迹系统是通过摄像头采集图像，所以对硬件的搭建的要求很严格，为达到降低误差的目的，要求各机械结构的尺寸设计精确，且装配准确。在设计循迹控制系统过程中，不仅仅提高了动手能力，更是提高了发现问题和解决问题的能力。PID 控制系统在硬件上还存在着误差，可进一步的通过设计的合理性及精确程度来降低误差，在软件上，单纯的使用 PID 控制器，今后可通过建立数学模型来提高控制的高效性和稳定性。

# 6 参考文献

# 7 附录

## 7.1 最小系统模块

此次设计，我们以 TivaTM4C123FH6PM 为主控芯片，自制主控制板。TivaTM4C123FH6PM 具有 2 个 PWM 波模块，可产生 16 路 PWM 波，为了防止多路 PWM 之间的相互干扰，PWM 的输入输出均采用光耦隔离电路，让输出和输入之间绝缘，提高电路的抗干扰性，信号传输更稳定。TivaTM4C123FH6PM 最多可提供 7 个 UART 串口，为了保证串口处电压安全，采用 TVS 管进行浪涌保护和过压保护。

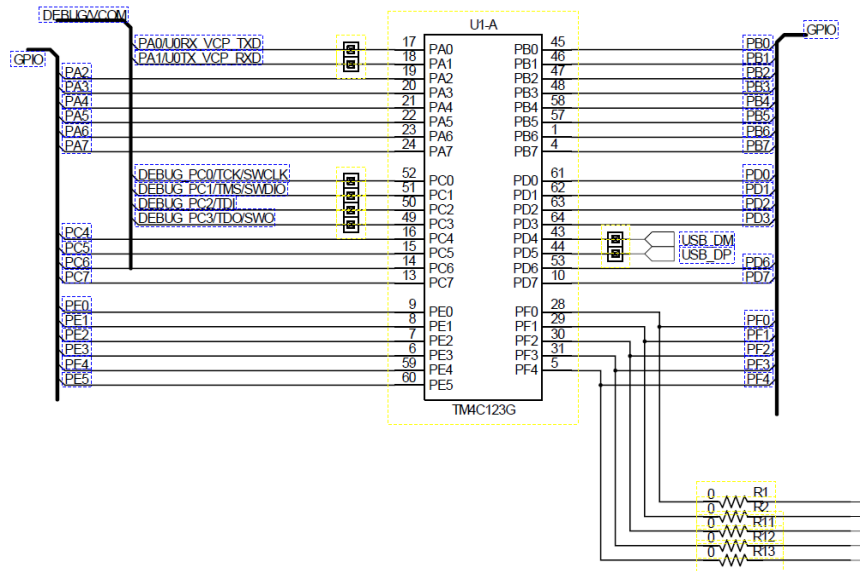

图 7-1

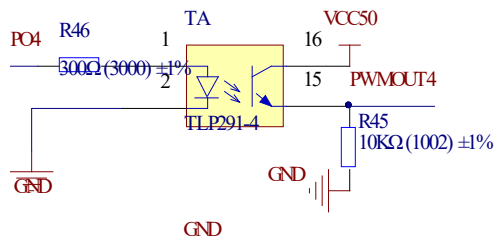

图 7-2

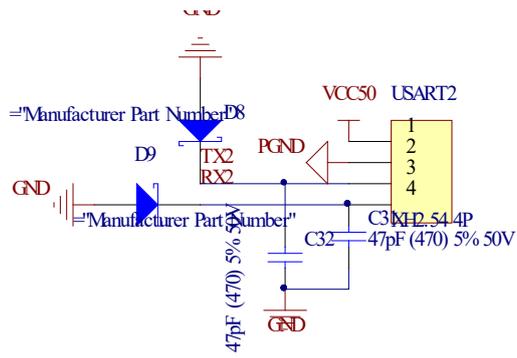

图 7-3

## 7.2 图像处理算法流程框图

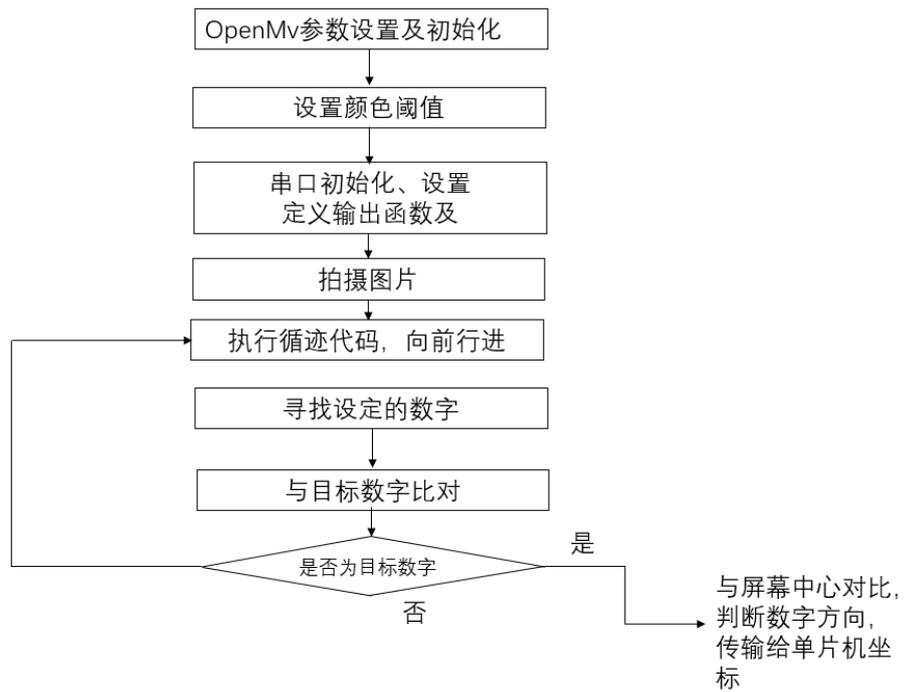

图 7-4